\documentclass[conference]{IEEEtran}
\IEEEoverridecommandlockouts
\usepackage{cite}
\usepackage{amsmath,amssymb,amsfonts}
\usepackage{algorithmic}
\usepackage{graphicx}
\usepackage{textcomp}
\usepackage{xcolor}
\usepackage{multirow}
\usepackage{dblfloatfix}
\usepackage{hyperref}
\usepackage{tikz}
\usepackage{subcaption}
\def\BibTeX{{\rm B\kern-.05em{\sc i\kern-.025em b}\kern-.08em
    T\kern-.1667em\lower.7ex\hbox{E}\kern-.125emX}}

\usepackage{etoolbox}

\begin{document}

\title{Predicting Chess Puzzle Difficulty with Transformers}

\author{\IEEEauthorblockN{1\textsuperscript{st} Szymon Miłosz}
\IEEEauthorblockA{\textit{Institute of Applied Computer Science} \\
\textit{Lodz University of Technology}\\
Lodz, Poland \\
szymonmilosz99@gmail.com}
\and
\IEEEauthorblockN{2\textsuperscript{nd} Paweł Kapusta}
\IEEEauthorblockA{\textit{Institute of Applied Computer Science} \\
\textit{Lodz University of Technology}\\
Lodz, Poland \\
pawel.kapusta@p.lodz.pl}
}

\maketitle
\begin{abstract}

This study addresses the challenge of quantifying chess puzzle difficulty - a complex task that combines elements of game theory and human cognition and underscores its critical role in effective chess training.
We present GlickFormer, a novel transformer-based architecture that predicts chess puzzle difficulty by approximating the Glicko-2 rating system. Unlike conventional chess engines that optimize for game outcomes, GlickFormer models human perception of tactical patterns and problem-solving complexity.

The proposed model utilizes a modified ChessFormer backbone for spatial feature extraction and incorporates temporal information via factorized transformer techniques. This approach enables the capture of both spatial chess piece arrangements and move sequences, effectively modeling spatio-temporal relationships relevant to difficulty assessment.

Experimental evaluation was conducted on a dataset of over 4 million chess puzzles. Results demonstrate GlickFormer's superior performance compared to the state-of-the-art ChessFormer baseline across multiple metrics. The algorithm's performance has also been recognized through its competitive results in the IEEE BigData 2024 Cup: Predicting Chess Puzzle Difficulty competition, where it placed 11th.

The insights gained from this study have implications for personalized chess training and broader applications in educational technology and cognitive modeling.

\end{abstract}

\begin{IEEEkeywords}
Chess puzzles, Transformer, Deep learning, Temporal modeling
\end{IEEEkeywords}
\section{Introduction}
Chess problems, or puzzles, have long been an essential part of chess training, designed to help players develop their tactical abilities by finding the best sequence of moves in a given position. These puzzles, derived from historic games or carefully crafted positions, can range from simple one-move solutions to complex multi-move combinations. However, assessing the difficulty of chess problems is not a straightforward task, as the factors influencing difficulty extend beyond basic game-theory principles. While the rules of chess and game objectives are well-defined, the complexity of a puzzle depends largely on human perception—players must identify patterns, recognize tactical motifs, and mentally navigate through a multitude of potential moves. This complexity underscores the importance of accurately evaluating puzzle difficulty levels, a task that is far from artificial. Indeed, it is crucial for chess players to engage with puzzles that match their skill level, ensuring they can effectively improve their abilities through appropriately challenging tasks. The need for such tailored training highlights why assessing puzzle difficulty is a significant and necessary aspect of chess education and improvement.

\begin{figure}[!t]
    \centering
    \includegraphics[width=\linewidth]{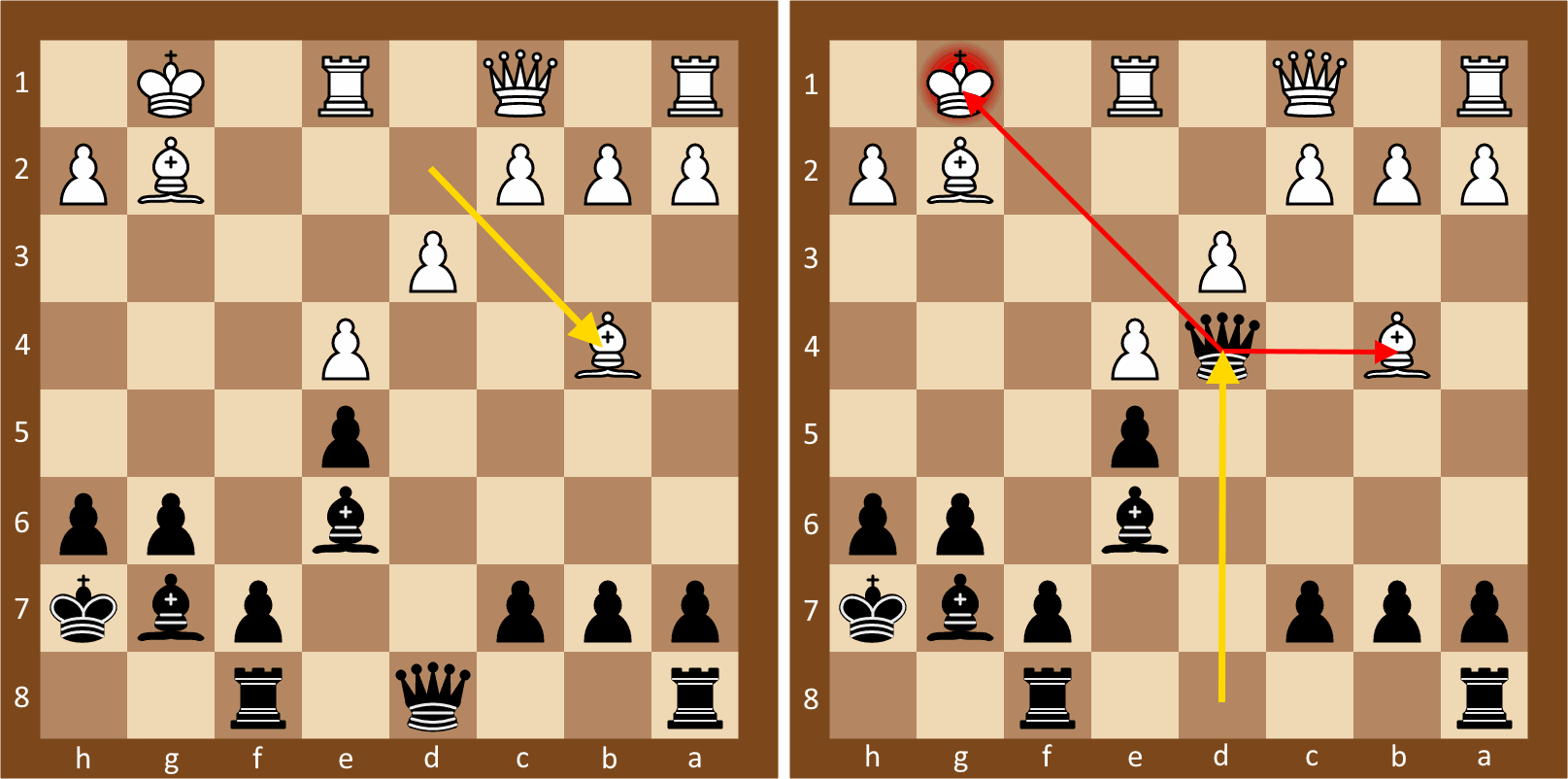}
    \caption{Example of a simple two-move chess puzzle; White played \textbf{Bb4} attacking Black's rook which is a blunder (left). The goal of the puzzle is to find the \textbf{Qd4+!} move, a fork attacking both the White king and bishop (right). In the next move White is forced to move the king or block the check, leaving the bishop undefended. Black then captures the unprotected bishop with \textbf{Qxb4}, resolving the puzzle by gaining material through the capture.}
    \label{fig:chess_puzzle_example}
\end{figure}

\begin{figure*}[!t]
    \centering
    \begin{subfigure}{0.48\linewidth}
        \includegraphics[width=\linewidth]{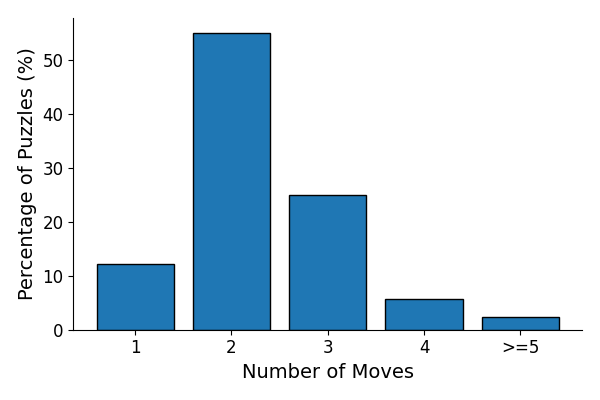}
        \caption{Distribution of the required number of moves to solve the puzzle in the dataset}
        \label{fig:chess_puzzle_length_distribution}
    \end{subfigure}
    \hfill
    \begin{subfigure}{0.48\linewidth}
        \includegraphics[width=\linewidth]{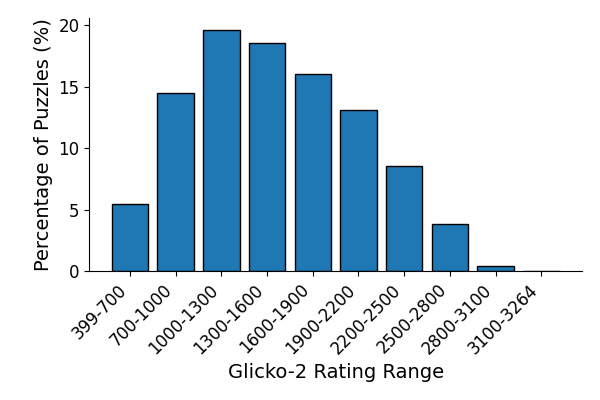}
        \caption{Distribution of puzzle difficulty in the dataset}
        \label{fig:chess_puzzle_diff_distribution}
    \end{subfigure}
    \caption{Distributions of selected dataset statistics}
    \label{fig:combined_puzzle_distributions}
\end{figure*}

Chess rating systems, such as Elo \cite{elo1978rating}, Glicko \cite{10.1111/1467-9876.00159}, and Glicko-2\cite{doi:10.1080/02664760120059219}, have long been used to quantify player strength. These systems provide a mathematical framework for estimating relative skill levels based on game outcomes. While these rating systems are well-established for player evaluation they offer insights into the challenge of assessing puzzle difficulty. However, this approach reveals a significant practical limitation: to obtain a reliable rating, each puzzle would need to be solved hundreds of times by players of varying skill levels. This resource-intensive process, while theoretically sound, proves impractical for large-scale puzzle evaluation or for assessing newly created challenges.

The task of predicting chess puzzle difficulty presents a unique challenge compared to traditional chess move prediction. In standard chess engines, the goal is to identify the move that maximizes the probability of winning, given a well-defined set of rules and a clear win condition. In contrast, assessing puzzle difficulty requires modeling the abstract concept of how players with varying skill levels perceive and solve tactical patterns.

Transformer architectures, which have revolutionized natural language processing and other domains, offer a promising approach to this challenge. Their ability to capture long-range dependencies and process sequential data makes them well-suited for analyzing chess positions and move sequences. Notably, the success of chess-specific transformers like ChessFormer \cite{monroe2024masteringchesstransformermodel}, utilized in the Leela Chess Zero chess engine, demonstrates the potential of these architectures in chess-related tasks.

In this paper, we present GlickFormer, a transformer-based model designed specifically for predicting chess puzzle difficulty by approximating the rating obtained with Glicko-2 system. By leveraging the power of attention mechanisms and the ability to capture complex patterns across both the spatial arrangement of chess pieces and the temporal sequence of moves, GlickFormer aims to provide a more accurate, consistent, and scalable method for assessing puzzle difficulty.

\section{Related Work}

Chess problem difficulty estimation is a relatively underexplored area in both artificial intelligence and chess research. Although considerable effort has been invested in developing chess engines capable of outperforming human grandmasters, much less attention has been given to understanding the cognitive challenges involved in solving chess puzzles. It is a non-trivial task to even define the difficulty of the problem. Previous work has shown, that assessing chess puzzle difficulty is hard even for human experts \cite{Hristova2014AssessingTD}. However, by following the suggestions of Grandmasters, some progress have been made on automatic problem diffculty assessment by mimicking the thought process of human players and constructing the meaningful search trees \cite{Stoiljkovikj2015ACM} \cite{10.1093/oso/9780198862536.003.0024}.

Chess engines like Stockfish and Leela Chess Zero (LCZero) have already demonstrated the power of game-tree search and neural networks to master chess at a superhuman level. LCZero, in particular, has shown evidence of a learned "look-ahead" strategy, as highlighted in recent research \cite{jenner2024evidencelearnedlookaheadchessplaying}. These models are trained to maximize winning probabilities and evaluate positions based on outcomes derived from large-scale self-play. However, GlickFormer diverges from these objectives. Instead of seeking optimal moves, it focuses on understanding the cognitive complexity behind problem-solving. In doing so, GlickFormer is designed to predict how difficult a puzzle will be based on how human players process information rather than how a machine would determine the best move.

The advent of transformers, as outlined in the "Attention is All You Need" paper \cite{vaswani2023attentionneed}, has revolutionized the field of artificial intelligence. Transformers excel at processing sequential data by employing a self-attention mechanism, which allows them to capture long-range dependencies and model intricate relationships in the data. This makes them particularly suited for tasks that involve both spatial and temporal components, such as video analysis or, in this case, chess puzzles. Recent advancements, such as the Video Vision Transformer (ViViT)\cite{DBLP:journals/corr/abs-2103-15691}, have demonstrated the ability of transformers to model complex patterns over time in non-Euclidean spaces\cite{feinashley2024hvtcomprehensivevisionframework}, making them an ideal choice for predicting the difficulty of chess puzzles.
Furthermore, a recent study \cite{ruoss2024grandmasterlevelchesssearch} has shown that large-scale transformer models trained on extensive chess datasets can achieve remarkable performance in chess, reaching a Lichess blitz Elo of 2895 against humans and solving challenging chess puzzles without explicit search algorithms. These developments underscore the potential of transformers in modeling complex game scenarios and predicting the difficulty of chess puzzles.

In a recent development, a new study has emerged \cite{omori2024chessratingestimationmoves}, which participated in the same IEEE BigData 2024 Cup competition \cite{organizers} as ours. It employs a combination of Convolutional Neural Networks (CNN) and bidirectional Long Short-Term Memory (LSTM) networks. While our approach leverages transformer architecture, this study opts for more traditional deep learning methods, while still incorporating temporal information. The CNN-LSTM approach achieved comparable, albeit lower, results on the public leaderboard in both preliminary and final scores, compared to our transformer-based solution. This parallel development in the field underscores the complexity of chess rating estimation and the diversity of potential approaches. While the CNN-LSTM model demonstrates the viability of traditional deep learning methods for this task, the superior performance of our transformer-based architecture suggests that the self-attention mechanism may be particularly well-suited for capturing the nuanced patterns in chess problem difficulty.

\section{Dataset}
The dataset consists of 4.2 million chess puzzles from Lichess.org, covering a wide range of difficulties from beginner to grandmaster levels (\autoref{fig:combined_puzzle_distributions}). Each puzzle includes:

\begin{itemize}
    \item The starting position in Forsyth-Edwards Notation (FEN).
    \item The sequence of moves leading to the solution.
    \item The Glicko-2 rating \( r \), indicating puzzle difficulty.
    \item The rating deviation \( RD \), reflecting assumed reliability in the rating.
    \item Additional metadata such as themes, number of plays, popularity scores, and associated game tags.
\end{itemize}

The mean and standard deviation of Glicko-2 rating values in the dataset are 1516 and 543 respectively. Majority of the Rating Deviation values in the dataset fall between 80 and 90, indicating a relatively high confidence in the difficulty ratings of these puzzles. However, puzzles that have been solved less frequently exhibit larger rating deviations, often exceeding 200. This higher RD reflects greater uncertainity in the estimated difficulty due to limited player interactions with the puzzles. 

\section{Methodology}
In this section, we describe our proposed approach for predicting chess puzzle difficulty. We begin with the puzzle encoding scheme that captures the spatial and temporal aspects of chess puzzles. We then detail our model architecture, which includes spatial feature extraction using the ChessFormer backbone and the incorporation of temporal information through two variants of our proposed model.

\subsection{Puzzle Encoding}
Chess puzzles are encoded as sequences of \( N \) boards, where each board is represented as \( B_n \in \mathbb{R}^{16 \times 8 \times 8} \). This encoding draws inspiration from AlphaZero's \cite{silver2017masteringchessshogiselfplay} approach, where similar board representation is used to capture the game state. The entire puzzle encoding is given by:
\[
\mathbf{P} = \{ B_1, B_2, \ldots, B_N \}.
\]
This encoding captures the intricate relationships between moves while maintaining a comprehensive representation of the chessboard and its pieces.

Each \( B_n \) is a binary tensor with 16 channels over an \( 8 \times 8 \) grid, organized as follows:

\begin{itemize}
    \item \textbf{Piece Representation (12 channels):}
    \begin{itemize}
        \item \textbf{6 Channels for Mover's Pieces:} Encoding the presence of each type of the mover's pieces (king, queen, rook, bishop, knight, and pawn). A value of 1 indicates the presence of the corresponding piece on a specific square, and 0 indicates absence.
        \item \textbf{6 Channels for Opponent's Pieces:} Similar encoding for the opponent's pieces, providing a consistent view from the mover's perspective.
    \end{itemize}
    \item \textbf{Move Encoding (4 channels):}
    \begin{itemize}
        \item \textbf{2 Channels for the Previous Move:} Encoding the starting and ending squares of the last move made, crucial for understanding the transition leading to the current state.
        \item \textbf{2 Channels for the Next Move:} Encoding the anticipated starting and ending squares of the next move, helping the model anticipate the upcoming state and contextualize the puzzle's solution.
    \end{itemize}
\end{itemize}
To focus on the solver's perspective, the board is mirrored if the mover is Black, ensuring consistency in representation.

The encoding process begins with the first position after the initial move that leads to the creation of the puzzle—often a blunder or inaccuracy. From this position, we encode every alternate position in the sequence:
\[
\mathbf{P} = \{ B_1, B_3, B_5, \ldots, B_N \}.
\]
By selecting every alternate position, we reduce redundancy while capturing essential changes that reflect the puzzle's evolution. Each \( B_n \) represents a specific state from the solver's perspective, ensuring the model can learn from decisions made throughout the puzzle-solving process.

\subsection{Model Architecture}

Our proposed model, named \textbf{GlickFormer}, is designed to predict the difficulty of chess puzzles as measured by their Glicko-2 ratings. GlickFormer effectively captures both spatial and temporal dynamics inherent in the puzzle-solving process. Drawing inspiration from the Vision Transformers for Video (ViViT) architecture \cite{DBLP:journals/corr/abs-2103-15691}, which addresses video classification by processing spatial and temporal information in a factorized manner, GlickFormer adapts these principles to handle sequences of chess positions. Each move in a chess puzzle is treated analogously to a frame in a video sequence.

\begin{figure}[!t]
    \centering
    \includegraphics[scale=0.87]{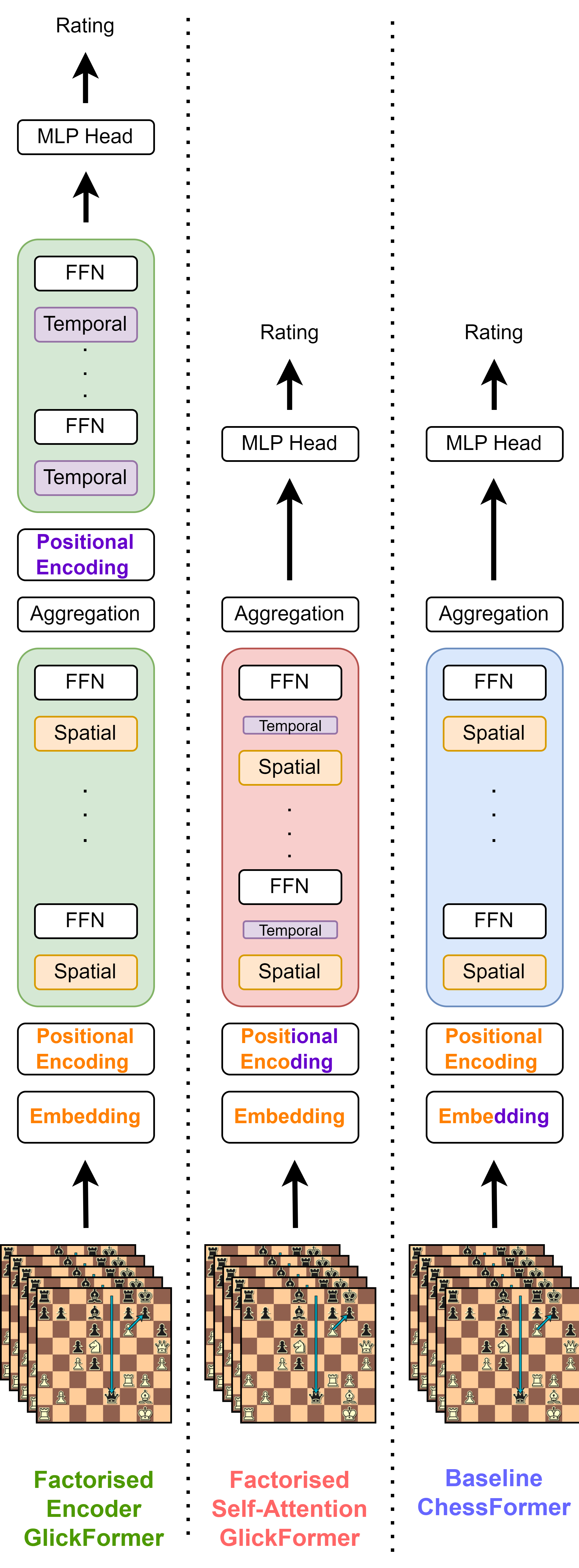}
    \caption{We propose two different architectures that incorporate temporal information processing mechanisms on top of the baseline ChessFormer model. Additionally, we modify the positional embedding schemes to address the architectural modifications.}
    \label{fig:model_architecture}
\end{figure}

We explore two variants of GlickFormer based on ViViT's factorization strategies: the \textit{Factorized Encoder GlickFormer} and the \textit{Factorized Self-Attention GlickFormer} (\autoref{fig:model_architecture}). The Factorized Encoder GlickFormer processes spatial features of each chess position using a transformer encoder and then incorporates temporal relationships via a separate temporal transformer encoder. In contrast, the Factorized Self-Attention GlickFormer integrates temporal information directly within the self-attention mechanism by factorizing the attention into spatial and temporal components within each transformer layer.

Both variants leverage the \textbf{ChessFormer} backbone \cite{monroe2024masteringchesstransformermodel} to extract rich spatial features from individual positions. Below, we detail the spatial feature extraction process and how temporal information is incorporated in each variant.

\subsubsection{Spatial Feature Extraction}

The spatial feature extraction consists of two main components: the ChessFormer backbone and the token transformation process. We employ the ChessFormer architecture as the backbone for our models due to its specialized design for processing chess positions.

Unlike traditional transformers that process tokens embedded in a topology whose structure roughly translates to Euclidean 1- and 2-space, ChessFormer adresses the unique topology of chess board, where the concept of distance is defined by chess-specific attributes rather than physical proximity. For example, squares that are a knight's, rook's or bishop's move away from each other are considered closely related even if they are far apart on the board. This is achieved by incorporating specialized positional encoding schemes within the self-attention mechanism, reflecting the inherent relationships between squares based on chess piece movement rather than Euclidean distance.

We utilize the variant of ChessFormer that introduces a learnable bias mechanism called \textit{Smolgen} \cite{leelaChessZeroBlogTransformerProgress} for each square in the self-attention mechanism before the softmax function. This approach allows the model to capture the unique characteristics and importance of each square on the chessboard, effectively reflecting the spatial relationships inherent in chess.

The bias matrix \( \mathbf{B} \) is critical in capturing the positional relationships between squares. For each block in the transformer encoder, it is obtained through a series of transformations applied to the token embeddings. Specifically, each token embedding \( \mathbf{h}_{i} \in \mathbb{R}^{d} \), representing the state of square \( i \), is first compressed to a lower-dimensional representation:
\[
\mathbf{c}_{i} = \mathbf{h}_{i} \mathbf{W}_c, \quad \mathbf{W}_c \in \mathbb{R}^{d \times d_c},
\]
where \( d_c = d / 32 \) is the reduced dimension. The compressed tokens are then concatenated and reshaped into a single vector:
\[
\mathbf{c} = \text{concat}(\mathbf{c}_{1}, \mathbf{c}_{2}, \ldots, \mathbf{c}_{64}) \in \mathbb{R}^{64 \cdot d_c}.
\]
This vector is passed through a fully connected layer with activation function \( \phi \) and layer normalization to obtain a hidden representation: 
\[
\mathbf{h} = \text{LayerNorm}(\phi(\mathbf{c} \mathbf{W}_h)), \quad \mathbf{W}_h \in \mathbb{R}^{(64 \cdot d_c) \times h_s},
\]
where \( h_s = d / 4 \) is the size of the hidden layer, and \( \phi \) denotes the activation function.

Subsequently, the hidden representation \( \mathbf{h} \) is transformed to generate bias embeddings for each attention head:
\[
\mathbf{g} = \text{LayerNorm}(\phi(\mathbf{h}_n \mathbf{W}_g)), \quad \mathbf{W}_g \in \mathbb{R}^{h_s \times (h \cdot h_s)},
\]
where \( h \) is the number of attention heads. The result \( \mathbf{g} \in \mathbb{R}^{h \cdot h_s} \) is reshaped into \( \mathbf{G} \in \mathbb{R}^{h \times h_s} \). A weight matrix \( \mathbf{W}_b \in \mathbb{R}^{h_s \times (64 \cdot 64)} \) shared between all encoder blocks is then used to compute the bias matrices for each attention head:
\[
\mathbf{b}^{(j)} = \mathbf{G}^{(j)} \mathbf{W}_b, \quad \mathbf{b}^{(j)} \in \mathbb{R}^{64 \cdot 64},
\]
where \( \mathbf{G}^{(j)} \) is the bias embedding for head \( j \). The bias vector \( \mathbf{b}^{(j)} \) is reshaped into the bias matrix:
\[
\mathbf{B}^{(j)} = \text{reshape}(\mathbf{b}^{(j)}, (64, 64)).
\]

For each attention head \( j \), the attention scores between tokens are computed as:
\[
\mathbf{A}^{(j)} = \frac{\mathbf{Q}^{(j)} (\mathbf{K}^{(j)})^\top}{\sqrt{d_k}} + \mathbf{B}^{(j)},
\]
where \( \mathbf{Q}^{(j)}, \mathbf{K}^{(j)} \in \mathbb{R}^{64 \times d_k} \) are the query and key matrices for head \( j \), and \( d_k = d / h \) is the dimensionality of each head.

The significance of this positional encoding lies in its ability to model the intricate positional dependencies between squares on the chessboard. In chess, the influence of a piece is not determined by Euclidean distance but by complex movement patterns unique to each piece type. Standard positional encodings used in transformers are inadequate for capturing these relationships. By utilization of Smolgen mechanism as described, the ChessFormer backbone can effectively model these complex interactions.

Using the ChessFormer backbone, we obtain the token embeddings \( \mathbf{H}_n \in \mathbb{R}^{64 \times d} \) for each board \( B_n \). To obtain a hidden representation for each board, we perform a token transformation process in a similar manner to how ChessFormer obtains value embeddings. Each token \( \mathbf{h}_{n,i} \) is linearly transformed to a lower-dimensional space:
\[
\mathbf{z}_{n,i} = \mathbf{h}_{n,i} \mathbf{W}_1 + \mathbf{b}_1, \quad \mathbf{W}_1 \in \mathbb{R}^{d \times d_z}.
\]
After transforming all tokens, we concatenate the resulting vectors:
\[
\mathbf{z}_n = \text{concat}(\mathbf{z}_{n,1}, \ldots, \mathbf{z}_{n,64}) \in \mathbb{R}^{64 \cdot d_z}.
\]
Finally, we project this concatenated vector to a higher-dimensional space using the transformation:
\[
\mathbf{s}_n = \mathbf{z}_n \mathbf{W}_2 + \mathbf{b}_2, \quad \mathbf{W}_2 \in \mathbb{R}^{(64 \cdot d_z) \times d_e}.
\]
The resulting vector \( \mathbf{s}_n \in \mathbb{R}^{d_e} \) serves as the embedding of board \( B_n \) for subsequent processing.

\subsubsection{Incorporating Temporal Information}

Before discussing our methods for incorporating temporal information, we must compare our approach to that of ChessFormer. ChessFormer integrates temporal dependencies by stacking several recent positions and embedding this temporal information directly into the tokens before processing them through the transformer encoder. In contrast, we process each position individually through the backbone and incorporate temporal dependencies at higher levels. This strategy allows us to capture complex temporal relationships.

We explore two methods for incorporating temporal information: the \textit{Factorized Encoder Model} and the \textit{Factorized Self-Attention Model}.

In the \textbf{Factorized Encoder Model}, we employ a late fusion strategy \cite{10.1145/1101149.1101236} to integrate spatial and temporal information. Initially, each board \(n\) is independently processed through ChessFormer encoder to obtain its representation \(s_n\). This approach allows the model to capture intricate spatial features of each board without the influence of temporal context. We construct a sequence matrix of these spatial representations:
\[
\mathbf{S} = [\mathbf{s}_1; \mathbf{s}_2; \ldots; \mathbf{s}_N] \in \mathbb{R}^{N \times d_e}.
\]
To incorporate temporal order and address the permutation invariance of self-attention mechanism, we add learnable positional encodings to each board representation:
\[
\tilde{\mathbf{s}}_n = \mathbf{s}_n + \mathbf{t}_n, \quad \mathbf{t}_n \in \mathbb{R}^{d_e}.
\]
Here, \(\mathbf{t}_n \) is a learnable embedding that encodes the position \(n\) of each board in the sequence, ensuring that the model can distinguish between different time steps.
We then process the sequence through \( L_t \) layers of transformer encoder blocks focusing on temporal relationships:
\[
\mathbf{S}^{(l)} = \text{TransformerLayer}^{(l)}_{\text{Temporal}}(\mathbf{S}^{(l-1)}), \quad l = 1, \ldots, L_t,
\]
where \( \mathbf{S}^{(0)} = [\tilde{\mathbf{s}}_1; \ldots; \tilde{\mathbf{s}}_N] \). The representation corresponding to the initial position \( \mathbf{s}_1^{(L_t)} \) is used to predict the puzzle's difficulty:
\[
\hat{y} = f(\mathbf{s}_1^{(L_t)}),
\]
with \( f: \mathbb{R}^{d_e} \rightarrow \mathbb{R} \) being a Multilayer Perceptron.

In the \textbf{Factorized Self-Attention Model}, we integrate temporal attention directly within the ChessFormer blocks by adding temporal attention layers after spatial attention. Each encoder block consists of spatial attention with Smolgen as described in \textit{paragraph 1)} of this subsection, temporal attention, and a feed-forward network (FFN). We apply residual connection and layer normalization after each of the aforementioned operation, which will be omitted from the following descriptions. Additionally we introduce a learnable embedding \( t_n \), which is added to each token embedding \(\mathbf{H}_n\) to encode the temporal position of each board \(n\) in the sequence of boards. This temporal positional encoding resolves the  self-attention invariance issue in temporal attention layers.

For each board \( n \), we apply ChessFormer's multi-head self-attention to obtain updated token embeddings \( \mathbf{H}_n^{(l,s)} \). We then stack the spatially attended tokens across all boards:
\[
\mathbf{H}^{(l,s)} = [\mathbf{H}_1^{(l,s)}; \mathbf{H}_2^{(l,s)}; \ldots; \mathbf{H}_N^{(l,s)}] \in \mathbb{R}^{N \times 64 \times d}.
\]
This tensor is reshaped to group tokens by spatial position:
\[
\mathbf{H}_{\text{reshaped}}^{(l,s)} \in \mathbb{R}^{64 \times N \times d}.
\]
For each spatial position \( i \), we extract the sequence of tokens across all boards:
\[
\mathbf{H}_{i}^{(l,s)} = [\mathbf{h}_{1,i}^{(l,s)}; \mathbf{h}_{2,i}^{(l,s)}; \ldots; \mathbf{h}_{N,i}^{(l,s)}] \in \mathbb{R}^{N \times d}.
\]
We apply multi-head self-attention over the temporal dimension:
\[
\mathbf{H}_{i}^{(l,t)} = \text{MultiHeadSelfAttention}(\mathbf{H}_{i}^{(l,s)}).
\]
After temporal attention, we reconstruct and reshape the tensor back to its original dimensions:
\[
\mathbf{H}_n^{(l,t)} = [\mathbf{h}_{n,1}^{(l,t)}; \mathbf{h}_{n,2}^{(l,t)}; \ldots; \mathbf{h}_{n,64}^{(l,t)}] \in \mathbb{R}^{64 \times d}.
\]
We then apply a position-wise FFN to each token:
\[
\mathbf{H}_n^{(l)} = \text{FFN}(\mathbf{H}_n^{(l,t)}).
\]
This process is repeated for \( L \) layers. After the final layer, we perform token transformation and aggregation to obtain the board representations \( \mathbf{s}_n \). The representation corresponding to the initial board \( \mathbf{s}_1 \) is used to predict the puzzle's difficulty:
\[
\hat{y} = f(\mathbf{s}_1).
\]

By integrating temporal information in these ways, GlickFormer effectively models the sequential nature of chess puzzles, capturing how the difficulty evolves with each move. The Factorized Encoder Model separates spatial and temporal processing, while the Factorized Self-Attention Model combines them within each layer, allowing us to explore different methods of capturing spatiotemporal dependencies.

\subsection{Training Objective}
\label{sec:Training_objective}
Both models are trained to minimize the mean squared error between the predicted difficulty \( \hat{y} \) and the ground truth difficulty \( y \):
\[
\mathcal{L} = \frac{1}{M} \sum_{i=1}^M (\hat{y}_i - y_i)^2,
\]
where \( M \) is the number of training samples in batch.

To improve the model's robustness and generalization, we incorporate uncertainty into the target labels by treating the standarized rating \( \mu_i = \frac{r_i - 1516}{543}\) and it's associated rating deviation \( \phi_i = \frac{RD_i}{543}\) as parameters of a normal distribution. Specifically, during training, instead of using the standarized Glicko-2 rating \( \mu_i \) directly as the target, we sample the target difficulty \( y_i \) from a normal distribution \( \mathcal{N}(\mu_i, \phi_i^2) \), effectively simulating various plausible difficulty values for each puzzle.

This approach serves two purposes:

\begin{itemize}
    \item \textbf{Data Augmentation:} By sampling from \( \mathcal{N}(\mu_i, \phi_i^2) \), we introduce variation in the target values for each puzzle, generating a broader range of difficulty ratings. This acts as a form of augmentation by providing the model with slightly different training targets for the same input puzzle, encouraging the model to learn general features that perform well across a variety of possible difficulty levels.
    
    \item \textbf{Regularization:} The introduction of variability in the target difficulty serves as a natural form of regularization. The model is encouraged to predict difficulty ratings that are stable across different samplings of the target, preventing it from overfitting to a specific Elo value and helping it generalize better to unseen data. Additionally, by clipping the sampled values to the range \( [\mu_i - 3\phi_i, \mu_i + 3\phi_i] \), we ensure that the sampled values remain within a reasonable range, preventing extreme outliers and further stabilizing training.
\end{itemize}

By combining these elements, the model is better equipped to handle the inherent uncertainty and variability in puzzle difficulty, resulting in improved performance and generalization.

\subsection{Implementation Details}
Following ChessFormer, we utilize the Mish activation function \cite{misra2020mishselfregularizednonmonotonic} 
\[
f(x) = x \tanh(\ln(1 + e^x)) 
\]
across all of our models. This choice is driven by Mish's smooth and non-monotonic properties, which have been shown to improve feature extraction and gradient flow in deep networks.\\
Key hyperparameters shared between all the models include:
\begin{itemize}
    \item Embedding dimension \( d = 256 \).
    \item Reduced dimesions \( d_c = d / 32 \) and \( d_z = 32 \) for token compression.
    \item Position embedding dimension \( d_e = 2d = 512 \).
    \item Number of encoder layers \( L, L_t = 16\).
    \item Number of self-attention heads \( h = 16 \).
    \item Dimensions of linear projections in self-attention heads \( d_k, d_q, d_v = d / h \).
    \item Maximum number of chessboard positions in the input sequence \(N_{max} = 5\).
\end{itemize}
For linear projections \( d_k, d_q, d_v\) in the temporal self-attention heads of Factorized Self-Attention GlickFormer we use a reduced dimension of \(d/2h\).

\section{Experiments}
\subsection{Experimental Setup}
We conducted experiments to evaluate the performance of GlickFormer in predicting chess puzzle difficulty. 
The dataset was split into training and validation sets, with 4,158,000 chess puzzles used for training and 42,000 puzzles reserved for testing and validation. The test set size, representing approximately 1\% of the total data, is sufficient to provide a reliable and statistically significant evaluation of the models' performance due to its substantial size and diversity. All models were implemented using Tensorflow 2.13 and trained on  Nvidia Quadro RTX 6000 GPUs.

We standarized the Glicko-2 ratings by subtracting the mean rating of 1516 and dividing by the standard deviation of 543 to facilitate training. During training, we sampled target ratings from a normal distribution centered at the standarized rating with a variance based on the rating deviation, as described in 
\autoref{sec:Training_objective}.

The models were trained for 28.000 steps using the \nobreak RMSprop optimizer with a learning rate of \(1 \times 10^{-6}\), \(\rho = 0.99\) and a batch size of 4096. We observed that momentum-based optimizers like Adam or higher learning rates caused the models to converge prematurely to predicting the mean of the dataset with zero standard deviation. Additionally, we employed a cyclical optimizer state restarting strategy {\cite{pokutta2020restartingalgorithmsfreelunch}}, where the length of the \(k\)-th cycle was set to \(1000k\) steps.

The standard distribution sampling was sufficient to prevent the models from overfitting, eliminating the need for additional regularization techniques such as dropout or weight decay.

\begin{figure}[t]
    \centering
    \includegraphics[width=\linewidth]{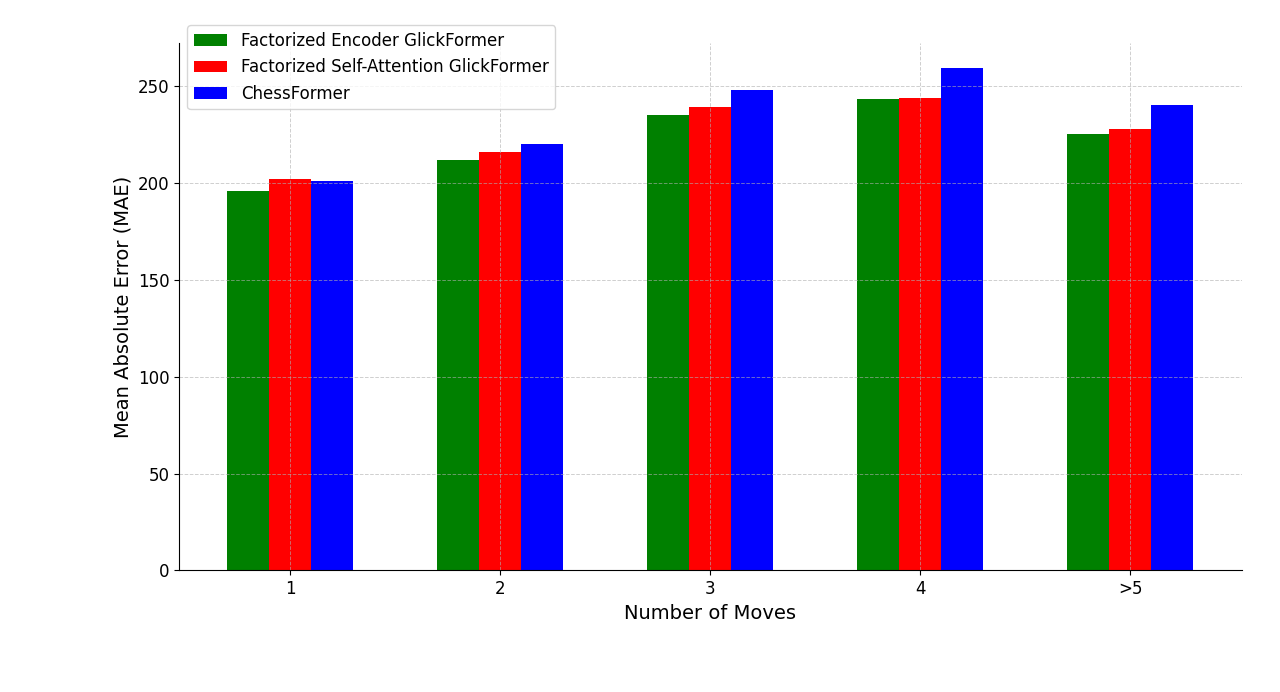}
    \caption{Mean Absolute Error of each model depending on number of moves required to solve the puzzle.}
    \label{fig}
\end{figure}

\subsection{Baseline Model}
To evaluate the effectiveness of our proposed models, we decided to compare it against a state-of-the-art solution. Therefore we have chosen a baseline model based on ChessFormer architecture. The model processes sequences of board states by stacking them along the input channels, following ChessFormer's method of incorporating temporal information by stacking the past \(N\) boards. This approach embeds temporal information directly into the input before it is processed by the transformer encoder. Furthemore, we utilize ChessFormer's value head to process the features extracted from the encoder in the same way as in our GlickFormer models. By comparing against this baseline, we can assess the impact of our proposed methods for modelling temporal dynamics on the performance of puzzle difficulty prediction.

\subsection{Results and Analysis}
We define three metrics to evaluate the performance of models on the test set. \textbf{Mean Absolute Error (MAE)}, \textbf{Mean Absolute Z-Score (MAZ)}, and \textbf{Accuracy within 1, 2, and 3 standard deviations}.

1. \textbf{Mean Absolute Error (MAE):}  
   The \textit{Mean Absolute Error} is given by the formula:
   \[
   \text{MAE} = \frac{1}{N} \sum_{i=1}^{N} \left| r_i - \hat{r}_i \right|
   \]
   where \(r_i\) represents the ground truth Glicko-2 ratings, \(\hat{r}_i\) are the predictions, and \(N\) is the total number of test samples. The MAE measures the average magnitude of errors in the predictions, indicating how far off the predictions are from the actual values on average.

2. \textbf{Mean Absolute Z-Score (MAZ):}  
   The \textit{Mean Absolute Z-Score (MAZ)} is calculated by normalizing the absolute error using the standard deviation of the ground truth values:
   \[
   \text{MAZ} = \frac{1}{N} \sum_{i=1}^{N} \left| \frac{r_i - \hat{r}_i}{RD_i} \right|
   \]
   where \(RD_i\) is the rating deviation assiociated with the ground truth rating value \(r_i\). The MAZ adjusts for the uncertainity of the ratings, allowing us to compare the magnitude of errors relative to the expected variation in the ground truth.

3. \textbf{Accuracy within 1, 2, and 3 Rating Deviations:}  

This metric measures the proportion of predictions that fall within 1, 2, and 3 standard deviations of the true values. In the context of the Glicko-2 rating system, the rating deviation (\( RD \)) reflects the uncertainty in a puzzle's difficulty rating, analogous to the standard deviation (\( \sigma \)) in a normal distribution.

For normally distributed data, the empirical rule states the following probabilities:

\begin{itemize}
    \item \textbf{68\%} of the data lies within 1 standard deviation (\( \sigma \)) of the mean.
    \item \textbf{95\%} of the data lies within 2 standard deviations (\( 2\sigma \)).
    \item \textbf{99.7\%} of the data lies within 3 standard deviations (\( 3\sigma \)).
\end{itemize}

This rule remains valid for the Glicko-2 system. For example, a player with a rating of 1500 and an \( RD \) of 50 has a real strength between 1400 and 1600 (two standard deviations from 1500) with 95\% confidence. Similarly, for puzzle difficulty ratings, the \( RD \) provides a confidence interval around the estimated difficulty.

The accuracy for each case is computed as:
\[
\text{Accuracy within } kRD = \frac{1}{N} \sum_{i=1}^{N} \mathbb{I} \left( |r_i - \hat{r}_i| \leq k RD_i \right)
\]
where \(k = 1, 2, 3\), and \(\mathbb{I}(\cdot)\) is the indicator function that returns 1 if the condition inside is true and 0 otherwise. This metric allows us to understand how many predictions fall within an acceptable range of variability from the true values.

\begin{table}[h]
\caption{Test Set Mean Absolute Error (MAE) and Mean Absolute Z-Score (MAZ)}
\label{tab:mae_maz}
\centering
\begin{tabular}{l c c}
\hline
\textbf{Model} & \textbf{MAE} & \textbf{MAZ} \\
\hline
ChessFormer & 227.00 & 2.68 \\
Factorized Self-Attention GlickFormer & 221.80 & 2.62 \\
Factorized Encoder GlickFormer & \textbf{217.71} & \textbf{2.57} \\
\hline
\end{tabular}
\end{table}

\begin{table}[h]
\caption{Test Set Accuracy within 1, 2, and 3 Rating Deviations}
\label{tab:accuracy}
\centering
\begin{tabular}{l ccc}
\hline
\multirow{2}{*}{\textbf{Model}} & \multicolumn{3}{c}{\textbf{Accuracy (\%)}} \\
 & \textbf{\textit{1RD}} & \textbf{\textit{2RD}} & \textbf{\textit{3RD}} \\
\hline
ChessFormer & 25.42 & 48.21 & 65.29 \\
Factorized Self-Attention GlickFormer & 26.13 &  48.54 & 66.08 \\
Factorized Encoder GlickFormer & \textbf{27.00} & \textbf{50.45} & \textbf{67.66} \\
\hline
\end{tabular}
\end{table}

The results of our experiments are presented in Tables \ref{tab:mae_maz} and \ref{tab:accuracy}. Our proposed models, Factorized Self-Attention GlickFormer and Factorized Encoder GlickoFormer, outperform the baseline ChessFormer model in all metrics.

The Mean Absolute Error (MAE) and Mean Absolute Z-Score (MAZ) are lower for our models, indicating more accurate predictions. Specifically, the Factorized Encoder GlickFormer achieves the lowest MAE of 217.71, demonstrating the effectiveness of separating spatial and temporal processing.

The accuracy within 1, 2, and 3 rating deviations also shows improvement with our models. The Factorized Encoder GlickFormer achieves the highest accuracy within all deviation ranges, with 67.66\% of predictions within 3 rating deviations of the true ratings.

Figure \ref{fig} depicts the distribution of MAE across puzzles in the test set, categorized by the number of moves required to solve them. This visualization highlights the enhanced performance of our models over the baseline, particularly in puzzles requiring multiple moves. However, an exception is observed with single-move puzzles, where the Factorized Self-Attention GlickFormer performed worse than the baseline.

These results confirm that incorporating temporal information significantly enhances the model's ability to predict puzzle difficulty. Modeling the sequence of moves allows the model to understand how the complexity of the puzzle evolves, leading to more accurate difficulty assessments.

Additionally we report the MSE on the test set of the IEEE BigData 2024 Cup: Predicting Chess Puzzle Difficulty competition. The prototype version of our Factorized Encoder model achieved a preliminary test set MSE score of 75,995 and a final MSE score of 158,292.

\section{Conclusion}
In this paper, we presented \textbf{GlickFormer}, a transformer-based model designed to predict the difficulty of chess puzzles by approximating the Glicko-2 rating system. By effectively capturing both the spatial and temporal dynamics inherent in chess puzzles, GlickFormer demonstrates the potential of transformer architectures in modeling complex human cognitive tasks beyond traditional gameplay.

Our approach leverages the ChessFormer backbone for spatial feature extraction and incorporates temporal information using methods inspired by prior work on factorized transformer architectures, specifically the \textit{Factorized Encoder Model} and the \textit{Factorized Self-Attention Model}. Through extensive experiments on a large dataset of over 4 million chess puzzles, we showed that GlickFormer outperforms the baseline state-of-the-art ChessFormer model across multiple metrics, including Mean Absolute Error, Mean Absolute Z-Score, and accuracy within 1, 2, and 3 rating deviations.

The results indicate that modeling the sequential nature of puzzles significantly enhances the ability to predict their difficulty, highlighting the importance of temporal information in understanding human problem-solving processes. Our models not only provide more accurate difficulty assessments but also demonstrate the applicability of transformer architectures in capturing abstract cognitive challenges, rather than merely optimizing for game outcomes.

This work contributes to the field by offering a new perspective on evaluating chess puzzle difficulty and underscores the potential of advanced deep learning models in educational and training contexts. Future work may include exploring more advanced temporal modeling techniques, such as incorporating attention mechanisms that focus on critical moves within a puzzle. Furthermore, integrating additional contextual information, such as player engagement data or thematic elements of puzzles, could further refine difficulty predictions. Extending this approach to other domains that require modeling human cognitive processes could provide valuable insights and applications.

\bibliographystyle{IEEEtran}
\bibliography{IEEEabrv,BigDataCup2024}

\end{document}